\documentclass{article}
\usepackage{spconf,amsmath,graphicx}

\usepackage{enumitem}
\setlist{nosep, leftmargin=14pt}

\usepackage{mwe} % to get dummy images
\usepackage{url}
\usepackage{hyperref}
\usepackage{cleveref}
\usepackage[dvipsnames]{xcolor}
\usepackage{pifont}
\newcommand{\cmark}{\textcolor{green!60!black}{\ding{51}}}
\newcommand{\xmark}{\textcolor{red!70!black}{\ding{55}}}

\title{CSEval: A Framework for Evaluating Clinical Semantics in Text-to-Image Generation}
\name{\begin{tabular}{c}
Robert Cronshaw, Konstantinos Vilouras, Junyu Yan, Yuning Du, Feng Chen\\
\textit{Steven McDonagh, Sotirios A. Tsaftaris}
\end{tabular}}

\address{School of Engineering, University of Edinburgh, United Kingdom}

\begin{document}
%\ninept
%
\maketitle
\begin{abstract}
Text-to-image generation has been increasingly applied in medical domains for various purposes such as data augmentation and education. Evaluating the quality and clinical reliability of these generated images is essential. However, existing methods mainly assess image realism or diversity, while failing to capture whether the generated images reflect the intended clinical semantics, such as anatomical location and pathology. In this study, we propose the Clinical Semantics Evaluator (CSEval), a framework that leverages language models to assess clinical semantic alignment between the generated images and their conditioning prompts. Our experiments show that CSEval identifies semantic inconsistencies overlooked by other metrics and correlates with expert judgment. CSEval provides a scalable and clinically meaningful complement to existing evaluation methods, supporting the safe adoption of generative models in healthcare. 
% Our code will be publicly available upon acceptance.
% The abstract should appear at the top of the left-hand column of text, about
% 0.5 inch (12 mm) below the title area and no more than 3.125 inches (80 mm) in
% length.  Leave a 0.5 inch (12 mm) space between the end of the abstract and the
% beginning of the main text.  The abstract should contain about 100 to 150
% words, and should be identical to the abstract text submitted electronically
% along with the paper cover sheet.  All manuscripts must be in English, printed
% in black ink.
\end{abstract}
\begin{keywords}
Generative Models, Text-to-image Generation, Evaluation Metrics, Clinical Semantics
\end{keywords}
%

% \begin{figure}
%     \centering
%     \includegraphics[width=1\linewidth]{}
%     \caption{Examples with high FID but low RadGraph--F1.}
%     \label{fig:opening}
% \end{figure}

\begin{figure}[t!]
\centering
\includegraphics[width=1\linewidth,keepaspectratio]{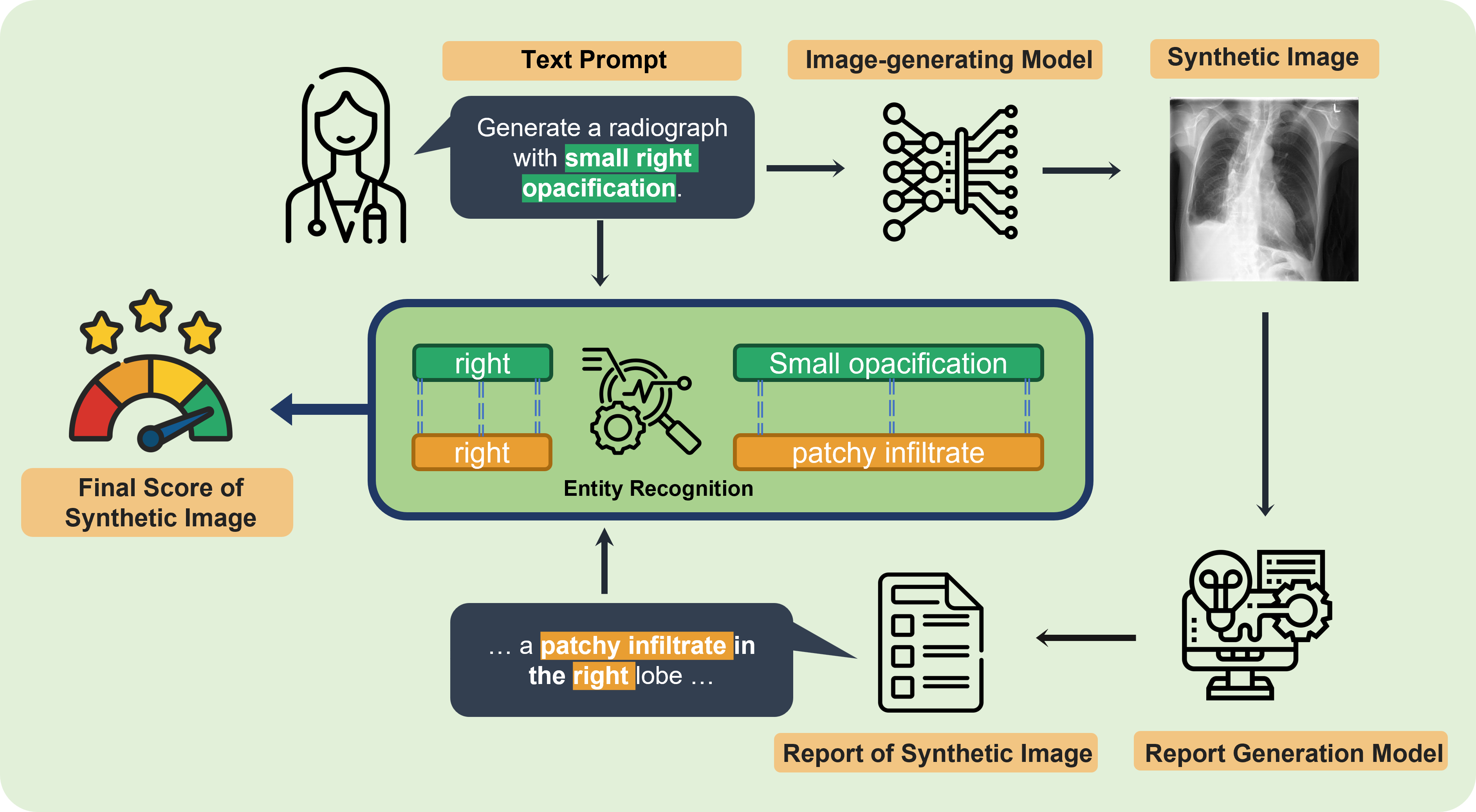}
    \caption{Our proposed framework for evaluating clinical semantics in text-to-image generation. A user-defined prompt guides the generation of synthetic medical images. Then, a pre-trained report generation model generates the findings of these synthetic images. The selected metric (RadGraph--F1 score), which measures the overlap between ground truth and synthetic entity-relation graphs in text space, quantitatively reflects how accurately the synthetic image adheres to the clinical details described in the original prompt.}
    \label{fig:framework}
\end{figure}

\begin{table*}[ht]
\centering
\resizebox{0.9\textwidth}{!}{\begin{tabular}{lcccc}
\hline
\textbf{Approach} & \textbf{Measure} & \textbf{Clinical Relevance} & \textbf{Image-prompt alignment} & \textbf{Scalability} \\
\hline
FID, MS-SSIM~\cite{heusel2016FID,wang2003msssim} & Distributional distance & \textcolor{Red}{Low} & \xmark & \cmark \\
CLIP/ BioViL-T~\cite{hessel2021clipscore} & Image--text alignment & \textcolor{Orange}{Medium} & \cmark & \cmark \\
Human Evaluation & Overall quality & \textcolor{Green}{High} & \cmark & \xmark \\
% CXR-specific~\cite{} & Classifier-based label detection & \textcolor{Orange}{Medium} & \cmark & \cmark \\
\textbf{CSEval (Ours)} & Clinical entity-aware & \textcolor{Green}{High} & \textbf{\cmark} & \textbf{\cmark} \\
\hline
\end{tabular}
}
\caption{Summary of evaluation metrics for synthetic images. Metrics are evaluated on three key criteria: clinical relevance (the capacity to capture clinical details such as pathology, severity and location), image-prompt alignment (the ability to measure adherence to the input prompt, but not necessarily related to clinical details), and scalability (the potential for automated, large-scale evaluation). Our proposed approach satisfies all three criteria.}
\label{tab:metrics}
\end{table*}

\section{Introduction}
\label{sec:intro}
The rapid advancement in Text-to-Image (T2I) generation models, such as Generative Adversarial Networks (GANs) and Diffusion Models, has revolutionised the synthesis of realistic and highly diverse visual content from natural language descriptions. These models are used to generate photorealistic images across broad areas including art and nature.

Recently, T2I generation has gained substantial attention in the medical domain \cite{khosravi2025exploring}. Its potential is actively explored to support critical applications such as data augmentation and medical education. Examples include the synthesis of detailed dermatological lesions and histopathology slides from simple descriptive prompts \cite{fayyad2025lesiongen, pielage2025interactive}. Therefore, evaluating the clinical semantic validity of these generated images is critical for the reliability and safe integration into medical workflows.

Current evaluation metrics such as Fréchet Inception Distance (FID) \cite{heusel2016FID}, Multi-Scale Structural Similarity Index (MS-SSIM) \cite{wang2003msssim}, and CLIP score \cite{hessel2021clipscore} are useful for quantifying visual fidelity, content diversity, and general semantic alignment, but are agnostic to clinical meaning \cite{deo2025metrics}. They measure the divergence between feature distributions or global similarity to a text prompt, but do not verify whether features in images, such as pathological findings, match the prompt in terms of anatomical site and severity. The insensitivity of existing metrics to clinically relevant semantic detail poses risks in deploying synthetic data that, while realistic, may contain clinically misleading content \cite{sun2023aligning}.

To address this limitation, we propose the Clinical Semantics Evaluator (CSEval) (Fig.\ref{fig:framework}) to assess the clinical fidelity (e.g., localisation and pathological correctness) of T2I outputs. Our CSEval reframes the evaluation process as a clinically meaningful task that mirrors human expert assessment by translating each generated image back into text. Specifically, a report generation module first produces a clinical description of the image. Then, an entity recognition module computes the F1 score to quantify the semantic alignment between the description and the original user prompt. We evaluate our framework on a dataset of 310 synthetic images and compare its performance with expert radiologist assessments. Our results show that CSEval successfully identifies semantic misalignments between prompt–image pairs that existing metrics fail to detect. We also summarise the main characteristics of existing metrics and CSEval in Table~\ref{tab:metrics}. Our main contributions are:
\begin{enumerate}
\item We propose CSEval, a framework for evaluating the clinical semantics of synthetic images to support scalable and reliable adoption of medical generative models.
\item CSEval features a modular and extensible design, integrating tailored components for report generation and entity extraction that can be further updated when more advanced models become available.
\item We demonstrate that CSEval effectively detects semantic misalignments between prompts and generated images that existing metrics fail to capture.
\item A domain expert (radiologist) was involved to validate CSEval’s assessments and ensure that the framework aligns with clinically relevant features. 
\end{enumerate}

\section{Related Work}
\label{sec:literature}

\subsection{Text-to-Image Generation in Medical Imaging}
Generative models such as Variational Autoencoders (VAEs) \cite{rais2024exploring}, Generative Adversarial Networks (GANs) \cite{singh2021medical}, and Diffusion Models \cite{kazerouni2023diffusion} have been widely applied to medical imaging tasks. More recently, the release of large image-text datasets (e.g., \cite{johnson2019mimic}) has enabled text-conditioned medical image synthesis at scale, with diffusion models achieving state-of-the-art results in various imaging modalities such as chest X-rays \cite{bluethgen2025vision,weber2023cascaded}, chest CT \cite{hamamci2024generatect,guo2025text2ct} and brain MR \cite{kim2024controllable}. Moreover, synthetic medical images must be both visually plausible and clinically faithful, accurately depicting the pathological findings described in the text prompt. However, the evaluation of such clinical fidelity remains limited.

\subsection{Evaluation Metrics for Synthetic Images}
Evaluating the quality of synthetic images has traditionally relied on quantitative metrics that assess image fidelity and diversity such as Fr\'echet Inception Distance (FID) \cite{heusel2016FID}, Inception Score (IS) \cite{salimans2016improvedtechniquestraininggans} and Multi-Scale Structural Similarity Index Measure (MS-SSIM) \cite{wang2003msssim}. However, since recent work has shown that these metrics do not align with human judgements \cite{stein2023exposing}, there has been a shift towards alternative evaluation paradigms more suitable for text-to-image generation \cite{hartwig2025survey}. These include training specialised models on large datasets of human ratings to learn preference scores \cite{xu2023imagereward} or prompting vision-language models (VLMs) to score image-text pairs \cite{huang2025t2i}. Such approaches, however, are not directly applicable to medical imaging: collecting human ratings at scale is prohibitively expensive, while general-domain VLMs are not suited for specialised medical images and domain-specific terminology. There also exist studies that assess the utility of synthetic medical images via downstream tasks \cite{deo2025metrics}, yet these evaluations ignore the text modality. 

\subsection{Clinical Evaluation Metrics for Automated Report Generation}
Focusing on the task of automated radiology report generation, common natural language generation metrics (e.g., BLEU and ROUGE) heavily penalise any lexical dissimilarities between real and synthetic text, while the underlying clinical semantics are not taken into consideration \cite{sloan2024automated}. To this end, there have been ongoing efforts to develop new metrics that correlate with human experts' preferences, leading to novel evaluation metrics that are either entity-based, such as RadGraph--F1 \cite{yu2023evaluating} and RaTEScore \cite{zhao2024ratescore}, or LLM-based such as GREEN score \cite{ostmeier2024green}. Inspired by recent developments in this field, we propose a framework that assesses text-to-image generators directly in the text space, thus focusing on the clinical attributes specified in the input text prompt.

\begin{table*}[ht!]
    \centering
    \resizebox{1.0\textwidth}{!}{\begin{tabular}{lcccc|c}
        \hline
        Finding & FID \( \downarrow \) & MS-SSIM (0–1) \( \downarrow \) & BioViL-T score (-1-1) \( \uparrow \) & \textbf{RadGraph--F1 (0–1) \( \uparrow \)} & Expert score (0–2) \( \uparrow \) \\
        \hline
        Cardiomegaly      & 7.42 & 0.503 $\pm$ 0.118 & 0.814 $\pm$ 0.098 & 0.528 $\pm$ 0.244  & 1.700 $\pm$ 0.174 \\
        Pleural effusion  & 6.57 & 0.431 $\pm$ 0.143 & 0.710 $\pm$ 0.192 & 0.163 $\pm$ 0.175 & 1.621 $\pm$ 0.164 \\
        Opacification     & 4.80 & 0.336 $\pm$ 0.145 & 0.556 $\pm$ 0.282 & 0.105 $\pm$ 0.143 & 0.920 $\pm$ 0.123\\
        Pneumothorax      & 8.08 & 0.425 $\pm$ 0.123 & 0.597 $\pm$ 0.142 & 0.043 $\pm$ 0.099 & 0.144 $\pm$ 0.064\\
        \hline
    \end{tabular}
    }
    \caption{Results on four thoracic diseases. We evaluate T2I generation quality using metrics (i) in the image space, such as FID (realism and diversity) and pairwise MS-SSIM (diversity); (ii) in a shared image-text space, such as BioViL-T score (alignment); (iii) in the text space, such as the RadGraph--F1 score proposed in this work; (iv) aligned with expert ratings.}
    \label{tab:results}
\end{table*}

\section{Methodology}
\label{sec:method}
\subsection{Main Idea}
We propose a novel framework for evaluating synthetic medical images, specifically designed to incorporate clinical semantic awareness. The proposed pipeline, as shown in Figure \ref{fig:framework}, consists of the following stages:
first, a user-defined prompt that provides a clear description of the underlying pathology, its anatomical location, and the severity level, is utilised to generate synthetic medical images via a Latent Diffusion Model (LDM) \cite{pinaya2022brain}. Second, a pre-trained report generation model (MAIRA-2) \cite{bannur2024maira} processes these synthetic images to produce corresponding clinical report findings. Third, an entity recognition model (RadGraph-XL) \cite{delbrouck2024radgraph} extracts clinical entities and their relations from both the original user prompts and the generated findings. Last, the F1-score, measuring the overlap between clinical entities extracted from the original and the generated text, respectively, serves as a quantitative measure that reflects the extent to which a synthetic image depicts the clinical attributes specified in the prompt.

\subsection{Design of CSEval}
\subsubsection{Prompt Template for Image Generation}
To systematically assess the quality of synthetic CXR scans in terms of their clinical content, we define a prompt template that covers a range of \textcolor{Red}{pathologies}, \textcolor{Green}{anatomical locations}, and levels of \textcolor{Blue}{severity}. As a result, we used the following prompts:
\begin{minipage}{\linewidth}
\begin{itemize}
    \item \texttt{\{\textcolor{Blue}{mild/moderate/severe}\} \textcolor{Red}{cardiomegaly}}
    \item \texttt{\{\textcolor{Blue}{small/moderate}\} \{\textcolor{Green}{left/right}\} \textcolor{Red}{pleural effusion}}
    \item \texttt{\{\textcolor{Blue}{small/moderate}\} \{\textcolor{Green}{left/right/left upper lobe/left lower lobe/right upper lobe/right lower lobe}\} \textcolor{Red}{opacification}}
    \item \texttt{\{\textcolor{Blue}{small/moderate/large}\} \{\textcolor{Green}{left/right/left apical/right apical}\} \textcolor{Red}{pneumothorax}}
\end{itemize}
\end{minipage}
where terms in \texttt{\{\}} denote options for a specific attribute. We also generate ten variations of synthetic scans with random initial noise for each prompt, resulting in 310 images in total.

\subsubsection{Report Generation Module}
To generate clinical free-text descriptions for synthetic images, we employ MAIRA-2 \cite{bannur2024maira}, a state-of-the-art grounded radiology report generation model. A key advantage of MAIRA-2 is its ability to provide spatial grounding for each described finding within the image through precise bounding box annotations. This inherent capability significantly enhances the quality and clinical relevance of the generated reports. Here, we retain only those sentences within the generated reports that contain explicit spatial localization which refer to a finding generated by MAIRA-2. %However, it is worth mentioning 
We note that our approach %can even 
also works with report generation models that can only output free-form text.

%This selective approach ensures that our analysis is specifically targeted at identifying clinical entities related to pathological conditions, anatomical anomalies, or varying levels of disease severity within the synthetic images.

\subsubsection{Entity Recognition Module}
RadGraph--XL \cite{delbrouck2024radgraph} is a domain-specific transformer-based entity recognition model that extracts clinical entities, such as observations and anatomical structures, and their relationships from radiology reports. In turn, the RadGraph--F1 score \cite{yu2023evaluating} compares the predicted entity-relation graph from synthetic reports against that of ground truth text (user-defined prompt). The score is defined as: 
% \begin{equation*}
%     \text{RadGraph--F1} = \frac{2 \cdot |G_{\text{pred}} \cap G_{\text{gt}}|}{|G_{\text{pred}}| + |G_{\text{gt}}|},
% \end{equation*}
\begin{align*}
P = \frac{|\mathcal{E}_{\text{pred}} \cap \mathcal{E}_{\text{gt}}|}{|\mathcal{E}_{\text{pred}}|}, \quad&
R = \frac{|\mathcal{E}_{\text{pred}} \cap \mathcal{E}_{\text{gt}}|}{|\mathcal{E}_{\text{gt}}|} \\
\text{RadGraph-F1} &= \frac{2PR}{P + R}
\end{align*}
where $\mathcal{E}_\text{pred}$ and $\mathcal{E}_\text{gt}$ denote the predicted and ground truth entity-relation tuples, respectively.

\section{Experiments}
\label{sec:exp}
\subsection{Implementation Details}
For the purposes of this study, we use publicly available models: the latent diffusion model\footnote{\url{https://github.com/Project-MONAI/GenerativeModels/tree/main/model-zoo/models/cxr_image_synthesis_latent_diffusion_model}} \cite{pinaya2022brain} was trained on MIMIC-CXR data \cite{johnson2019mimic}; MAIRA-2\footnote{\url{https://huggingface.co/microsoft/maira-2}} \cite{bannur2024maira} used for report generation is trained on a mix of public and private data, including MIMIC-CXR; and RadGraph-XL\footnote{\url{https://github.com/Stanford-AIMI/radgraph}} \cite{delbrouck2024radgraph} used for evaluation is also trained on MIMIC-CXR and a private CXR dataset.

\begin{figure}[t!]
    \centering
    \includegraphics[width=0.8\linewidth,keepaspectratio]{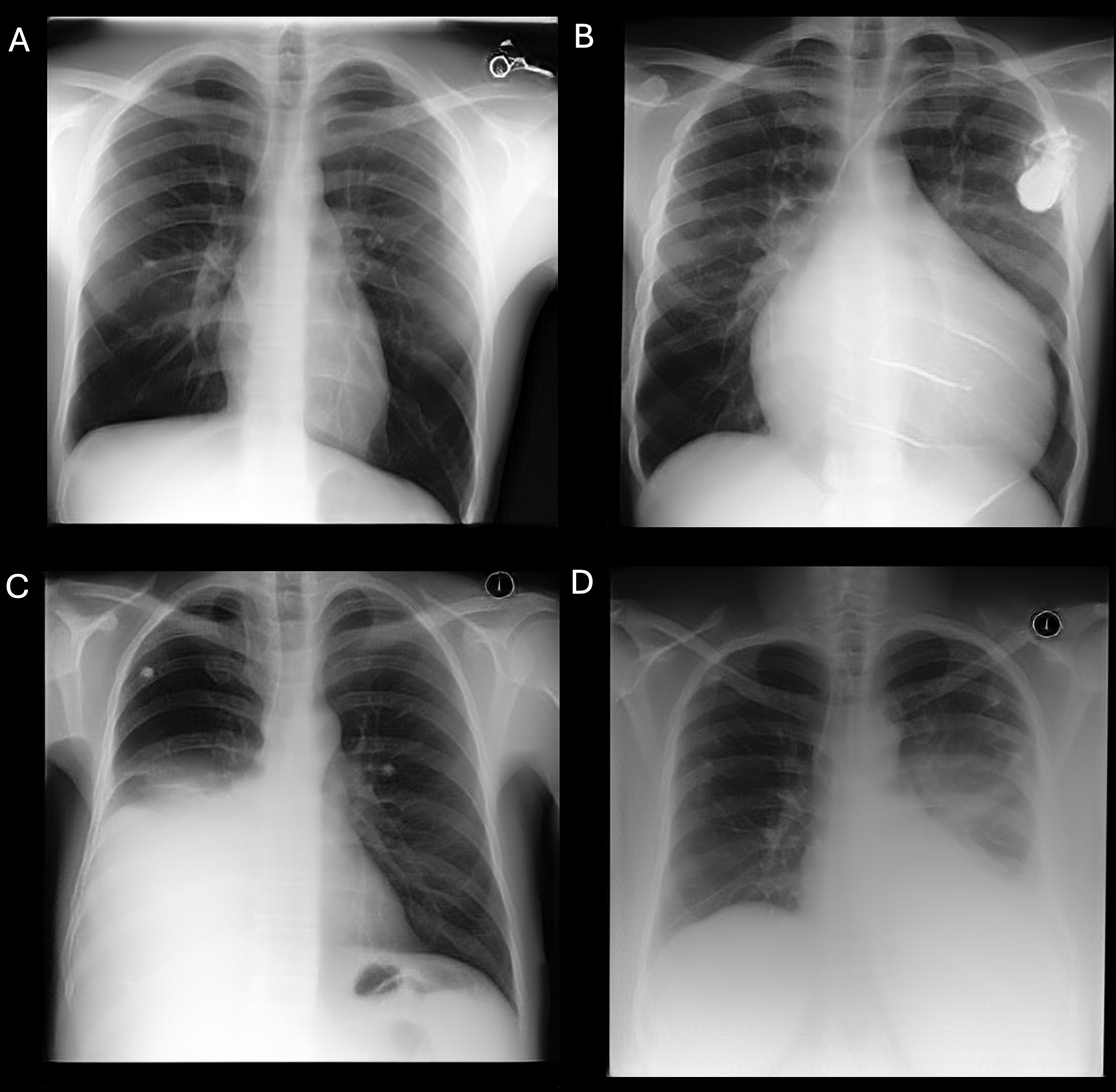}
    \caption{Synthetic examples with user prompts and RadGraph--F1 scores. A: large left apical pneumothorax (0.00), B: severe cardiomegaly (0.67), C: moderate right pleural effusion (0.29), D: small left lower lobe opacification (0.20).}
    \label{fig:exampleimgs}
\end{figure}

\subsection{Results}
We compared our proposed metric, termed CSEval (using RadGraph-F1), with FID, MS-SSIM, and BioViL-T (CLIP) score \cite{bannur2023learning}. The expert scores were provided by a domain expert (Radiologist with four years experience and board certification equivalent qualification [FRCR]), who rated the extent to which generated images aligned with the prompt with regards to disease type, severity and location, where applicable. They were summarised on a three point scale according to semantic alignment between prompt and image (0 = no alignment, 1 = some alignment --- correct pathology but wrong severity/location, 2 = strong alignment). The rater was unblinded to the disease label for the image being rated. A global assessment of image fidelity was not included in the score.
% \begin{table}
%     \centering
%     \begin{tabular}{ccccll}\toprule
%          &  FID &  MS-SSIM (0 -1)& CLIP&RadGraph--F1 (0 -1) &Expert score (0 - 2)  \\\midrule
%          Cardiomegaly&  7.42&  0.503& 0.814&0.524&1.94  \\
%          Pleural effusion&  6.56&  0.431& 0.710&0.177&1.63  \\
%          Opacification&  4.80&  0.336& 0.556&0.103&0.920  \\
%          Pneumothorax&  8.08&  0.425& 0.596&0.042&0.144  \\
%  & & & & &\\
%  Spearman correlation coefficient& -0.2& 0.8& 0.8& 1&\\
%  p-value& 0.8& 0.2& 0.2& 0&\\ \bottomrule
%     \end{tabular}
%     \caption{Caption}
%     \label{tab:results}
% \end{table}
%We compared each quality metric with the `ground-truth' expert %rating score according to Spearman's rank, (Table 1). With %regards to the expert scoring, FID was not at all correlated ($\rho = -0.2$, p = 0.8), MS-SSIM and BioViL-T CLIP were correlated but not statistically significantly ($\rho$ = 0.8, p = 0.2) and RadGraph--F1 was highly correlated ($\rho$ = 1, p $<$ 0.005).

As shown in Table~\ref{tab:results}, existing metrics often fail to reflect clinical semantics. For example, MS-SSIM and BioViL-T scores for \textit{Pneumothorax} are comparable to those of \textit{Opacification}, yet expert scores reveal a significant performance gap (0.144 vs.\ 0.920). Similarly, FID penalises \textit{Cardiomegaly} (7.42) as severely as \textit{Pneumothorax} (8.08), contrary to expert scores. In contrast, our proposed CSEval score (RadGraph-F1) is the only metric whose rank ordering of pathologies is consistent with the expert's assessment, demonstrating its superior ability to capture clinical semantics. Moreover, Kendall's $\tau$ rank correlation between the per-image metrics (BioViL-T and ours) and the expert scores shows that our method has a stronger correlation ($\tau$ = 0.375) than BioViL-T score ($\tau$ = 0.291). This result confirms that CSEval aligns more closely with expert judgment. 
%Note that distribution-level metrics (FID and MS-SSIM) were excluded from the per-image correlation analysis.

Qualitative examples in Fig.~\ref{fig:exampleimgs} visually support these quantitative findings. The figure displays generated images, their conditioning prompts, and corresponding CSEval scores, illustrating our method's ability to capture varying degrees of clinical-semantics alignment.

\section{Conclusion}
\label{sec:conclusion}
This work underscores the critical need of integrating clinical semantics into the comprehensive evaluation of synthetic medical images. We introduced CSEval, a novel framework that effectively measures the clinical relevance and semantic alignment between an image and its conditioning prompt. Our proposed metric correlates with expert judgements, proving its ability to capture clinical semantics. Furthermore, the framework is scalable and highly extensible, allowing its modular report generation and entity recognition components to be readily updated. We acknowledge two primary limitations. First, the performance of CSEval depends on the underlying report generation model's accuracy. While we mitigate this by employing a state-of-the-art model, errors in report generation can propagate to the final evaluation score. Second, the RadGraph-F1 proxy for semantic similarity can be influenced by textual artifacts (e.g., formatting, sentence length) that are distinct from pure clinical content, potentially introducing noise. Future work should focus on developing refined methodologies for calculating entity similarity, aiming to address these nuances and further enhance CSEval's precision and interpretability.

\section{Compliance with ethical standards}
\label{sec:ethics}
This research study was conducted retrospectively using human subject data made available in credentialed access via MIMIC-IV. Access was granted with appropriate credentialing under the terms of the PhysioNet Credentialed Health Data License and Health Data Use Agreement 1.5.0.  

\section{Acknowledgements}
\label{sec:acknoledgement}
This work was supported by the Wellcome Trust 340156/Z/25/Z

\bibliographystyle{IEEEbib}
\bibliography{strings,refs}

\end{document}